\documentclass[10pt,twocolumn,letterpaper]{article}

\usepackage{dicta}
\usepackage{times}
\usepackage{epsfig}
\usepackage{graphicx}
\usepackage{amsmath}
\usepackage{amssymb}
\usepackage{amsfonts}
\usepackage{mathrsfs}
\usepackage{makecell}

\usepackage[pagebackref=true,breaklinks=true,letterpaper=true,colorlinks,bookmarks=false]{hyperref}

\dictafinalcopy % *** Uncomment this line for the final submission

\def\dictaPaperID{163} % *** Enter the DICTA Paper ID here

\ifdictafinal\fi
\begin{document}

%%%%%%%%% TITLE
\title{Enhancing Fine-Grained Visual Recognition in the Low-Data Regime Through Feature Magnitude Regularization}

\author{Avraham Chapman*,
Haiming Xu$^\dagger$ and Lingqiao Liu$^\ddagger$\\
The University of Adelaide\\
Adelaide, Australia\\
Email: *avraham.chapman@adelaide.edu.au,
$^\dagger$hai-ming.xu@adelaide.edu.au,
$^\ddagger$lingqiao.liu@adelaide.edu.au}

\maketitle
%\thispagestyle{empty}

%%%%%%%%% ABSTRACT
\begin{abstract}
Training a fine-grained image recognition model with limited data presents a significant challenge, as the subtle differences between categories may not be easily discernible amidst distracting noise patterns. One commonly employed strategy is to leverage pretrained neural networks, which can generate effective feature representations for constructing an image classification model with a restricted dataset. However, these pretrained neural networks are typically trained for different tasks than the fine-grained visual recognition (FGVR) task at hand, which can lead to the extraction of less relevant features. Moreover, in the context of building FGVR models with limited data, these irrelevant features can dominate the training process, overshadowing more useful, generalizable discriminative features.
Our research has identified a surprisingly simple solution to this challenge: we introduce a regularization technique to ensure that the magnitudes of the extracted features are evenly distributed. This regularization is achieved by maximizing the uniformity of feature magnitude distribution, measured through the entropy of the normalized features. The motivation behind this regularization is to remove bias in feature magnitudes from pretrained models, where some features may be more prominent and, consequently, more likely to be used for classification. Additionally, we have developed a dynamic weighting mechanism to adjust the strength of this regularization throughout the learning process. Despite its apparent simplicity, our approach has demonstrated significant performance improvements across various fine-grained visual recognition datasets.
\end{abstract}

\section{Introduction}
\label{sec:intro}

\begin{figure}
\centering
    \includegraphics[width=3.2in]{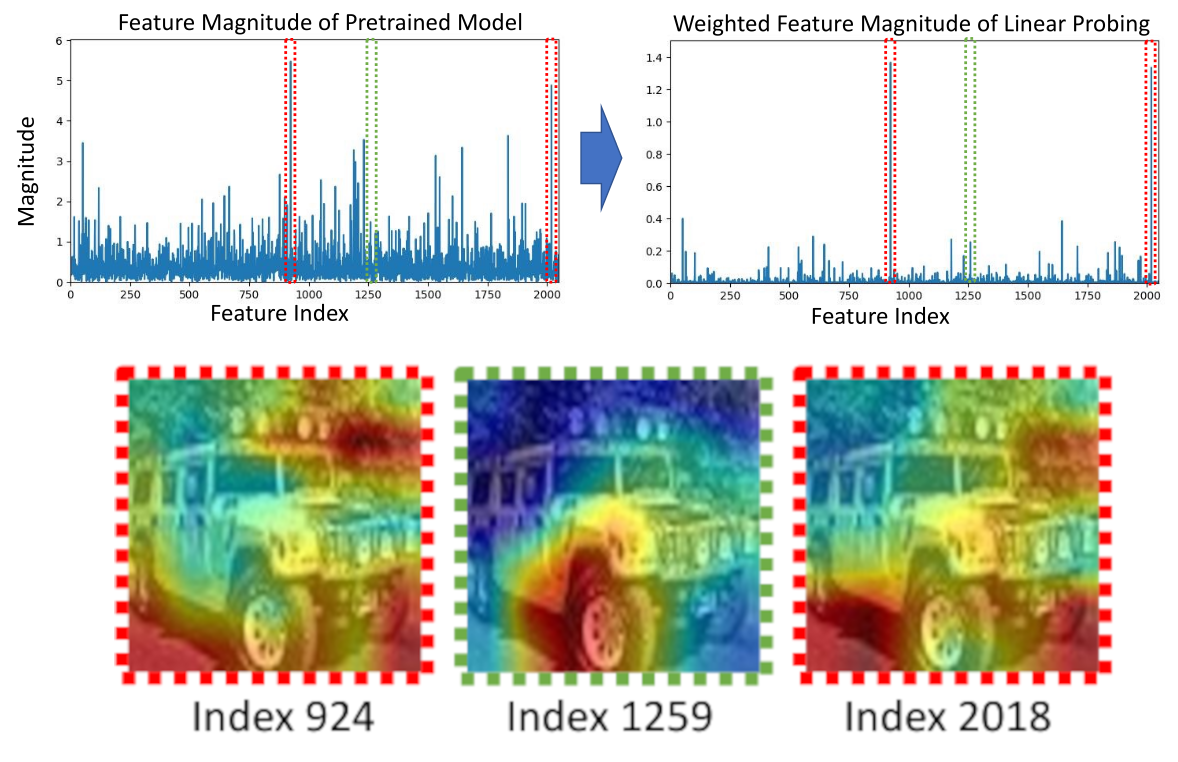}
    \caption{Pre-existing bias in a training dataset can lead to classifiers focusing on features that are not important, often to the detriment of useful features. See more explanation in Section~\ref{subsec:feat-mag-bias}.}
    %\hm{Question: how to define what is good or bad feature?}}
    \label{fig:feature_weighting_problem}
\end{figure}

Fine-grained visual recognition (FGVR) involves the classification of a large number of groups that differ only subtly from each other. Differentiating these classes often requires sensitivity to specific features in small regions of an image. For a bird, the difference between the two species may lie in the subtle differences between their beaks or feather-tips~\cite{CUB200Birds}. Training a model to discover these features is further complicated by the fact that there is often a dearth of data available for training. The more specialized the dataset, the more difficult it is to find the expertise to label the images~\cite{Sukhbaatar2015, TongXiao2015}. A model trained on a limited FGVR dataset can often be sidetracked by irrelevant details, such as background features in an image.

Many foundational vision models exist and have demonstrated success across various downstream tasks~\cite{DINO, MocoV2, SimCLR, BYOL}. When applied to FGVR tasks, these models can deliver reasonable results even though they are not specifically tailored for FGVR~\cite{SAMBilinear,Wang2021,Kim2022}. The challenge arises because these models, while robust, are not optimized for FGVR's unique requirements. Consequently, when these models are fine-tuned using a limited number of labeled samples for FGVR, there is a risk that the most transferable and distinctive features crucial for FGVR may not be effectively emphasized. Additionally, there is a concern that these models may inherit bias from their pretraining phase. One particular form of bias manifests in the feature magnitudes, where certain dimensions of the feature space are more likely to exhibit significant values. Consequently, these dimensions are more likely to be utilized if they exhibit discriminative patterns within the training dataset. However, when dealing with a small training dataset, the identified discriminative features may not generalize well to unseen test images. For instance, these features may overly focus on background regions, which is not conducive to accurate FGVR, as presented in Figure~\ref{fig:feature_weighting_problem}.

In this study, we propose a compellingly straightforward approach to enhance fine-grained image recognition when working with sparse data. Our method introduces a regularization strategy known as Feature Magnitude Regularization (FMR), which aims to equalize the distribution of feature magnitudes across the model.
By computing the entropy of normalized features and striving to maximize this entropy, we ensure a more balanced feature representation. This approach is specifically designed to encourage an equitable importance among all features during the training process, thus mitigating potential biases in feature magnitudes. An important consideration when applying this regularization is how to adjust its strength effectively.
Instead of employing a fixed weight for the regularization, we have developed a dynamic weighting mechanism that adapts the strength of regularization as the learning process unfolds. To achieve this, we set the regularization strength in proportion to the disparity between the current entropy of feature magnitude distribution and its maximum value. This encourages stronger regularization when the feature magnitude distribution deviates significantly from uniformity, ensuring that our approach remains effective throughout the whole training procedure. 

We performed extensive experimental evaluations on several popular fine-grained visual recognition benchmarks. Our experiments clearly demonstrate that the proposed method yields substantial improvements over conventional fine-tuning techniques when working with limited data. Furthermore, our approach exhibits favorable performance compared to other methods specifically designed to enhance the fine-tuning of pretrained models with a limited amount of data.

\section{Related Works}
\label{sec:relatedworks}

FGVR is concerned with the classification of multiple fine-subcategories of a larger group. There have been attempts to address this problem as far back as 25 years ago \cite{JOHNSON1998515, CUB200Birds}. The advent of deep learning \cite{DeepLearning} provided a powerful tool to address this problem.

Approaches tend to be grouped into the following two areas: Recognition by Localization-Classification Sub-networks~\cite{shroff2020focus,wang2018learning} and Recognition by End-to-End Feature Encoding~\cite{liu2014treasure, Zeiler2011AdaptiveDN,lin2017bilinear}.

Recognition by Localization-Classification Sub-networks work~\cite{shroff2020focus,wang2018learning} by attempting to locate key parts of an image, such as a bird's beak, and extracting feature vectors describing each part. These feature vectors, along with feature vectors describing global aspects of the image, are then passed to sub-networks, whose job is to perform classification. Examples include R-CNN \cite{RCNN}, FCN \cite{FCN}, and Faster R-CNN \cite{FasterRCNN}. Another more recent example is SAM-Bilinear~\cite{SAMBilinear}, which uses a self-boosting mechanism to build up an understanding of which regions of an image are relevant for the FGVR task.

Recognition by End-to-End Feature Encoding is about guiding convolutional neural networks (CNNs) to learn features from an input that provides enough discriminative information to allow for distinguishing subtle differences between similar classes. Methods of achieving this include higher-order feature interactions and novel loss functions. Higher-order feature interaction-based methods involve mining higher-order feature statistics from deeper convolution layers to extract useful descriptions of object parts~\cite{liu2014treasure, Zeiler2011AdaptiveDN}. Bilinear Convolutional Neural Networks (B-CNNs)~\cite{lin2017bilinear} use two CNNs whose outputs at each location combined to form a bilinear feature representation.

Another popular method for boosting FGVR performance is through the introduction of loss functions. These functions may attempt to reduce the confidence of predictions by the model~\cite{dubey2018pairwise} or to learn correlations between feature regions~\cite{gao2020channel, zhuang2020learning}. In addition, there are techniques like MC-Loss~\cite{MCLoss}, which attempt to locate harder classes and boost their gradients to encourage learning of the harder classes. Finally, there are losses that attempt to do a better job exploiting the knowledge already contained in a pre-trained model. L\textsuperscript{2}-SP \cite{L2Sp} uses a simple L2 penalty to encourage similarity between the final weights after target dataset training and the initial weights before training. DELTA \cite{li2020delta} encourages a similarity between the output of the encoder before and after training on carefully selected features using channel-wise attention. Batch Spectral Shrinkage (BSS)~\cite{BSS} attempts to avoid negative transfer by suppressing smaller singular value components. Co-Tuning \cite{CoTuning} sets out to establish the relationship between the source dataset classes and the target dataset classes, converting one-hot vectors across the logits for one dataset to probability distributions across the logits for the other dataset. It then trains both tasks in tandem. MaxEnt \cite{dubey2018bentropy} used Kullback–Leibler divergence to encourage the entropy across the logits to be as high as possible, thus reducing unwarranted confidence in the classifier.

These loss-based techniques are the most similar to our own work and will form a basis of comparison in Section~\ref{sec:experiments}.

\section{Method}
\label{sec:method}

% This section describes in detail the proposed method. The notations and definitions used in this paper are as follows:

% \begin{itemize}
%   \item $N$: The total number of samples being trained on.
%   \item $C$: The total number of classes in the dataset.
%   \item $B$: The total number of samples in each mini-batch.
%   \item $\Psi$: The Feature Extractor.
%   \item $D$: The dimension of the feature vector output from $\Psi$.
%   \item $h_\text{cls}$: A linear layer for classification.
%   \item $\mathscr{L}_\text{cls}$: The classification task loss.
%   \item $\mathscr{L}_\text{fmr}$: The FMR task loss.
%   \item $\lambda$: A weighting for the FMR loss.
%   \item $\beta$: The desired initial value for $\lambda$ to be provided by Dynamic Coefficient Tuning.
% \end{itemize}

\subsection{Method Overview}
Our method tackles the challenge of training a fine-grained image classification model when the available dataset has a limited number of training samples.

%This can lead to the selection of inappropriate features for classification that fail to generalize to the population as a whole. These features serve as distractions to the classifier. We aim to encourage the learning of useful features that will generalize well. While all learned features will high discriminative value on the training set, only the useful features will have high discriminative value with the test set. On the other hand, a distracting feature would have high discriminative value on the training set and little discriminative value on the test set. This observation gives use a way of quantifying the number of useful vs distracting features learned. Subsection \ref{subsec:encouraging_useful_features} uses this approach to show that FMR increases the number of useful features.

\begin{figure}
\centering
    \includegraphics[width=3in]{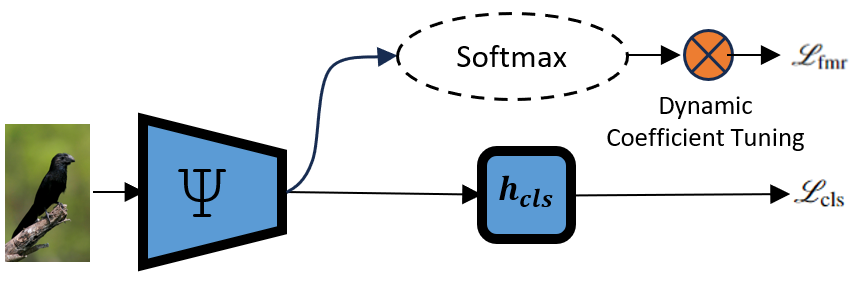}
    \caption{Our approach introduces an auxiliary task that encourages the magnitudes of the feature vectors to have less variability. The weighting of the task is dynamically set during training.}
    \label{fig:fmr_model}
\end{figure}

The overall structure of our technique is depicted in Figure \ref{fig:fmr_model}. As illustrated, our approach involves the introduction of an extra loss term alongside the standard cross-entropy loss typically used in supervised learning. After extracting features from the network backbone $\Psi$, we begin by applying softmax normalization to these features, transforming them into a representation resembling a probability distribution. Subsequently, we calculate the negative entropy of this distribution-like representation and employ it as a form of regularization loss, using a dynamically calculated weighting. 
The following sub-sections describe our proposed network and go into more detail about the feature magnitude regularization (FMR) training process.

\subsection{Feature Magnitude Bias}
\label{subsec:feat-mag-bias}
%Using pretrained models is a common practice to build image classification systems with a small amount of training data. Indeed, the pretrained models offer high-quality feature representations and can well capture the visual content of an image. However, on the other side, the pretrained models are often trained from an image dataset that are quite different from the downstream fine-grained recognition task. This could potentially introduce some bias in the feature representations, that is, some visual content are better captured in the resulted features. However, those visual content might not be necessarily useful for the downstream task. One may expect that those features will not be used by the classifier learned from the downstream task training data or will be suppressed during training. However, our preliminary investigation suggests otherwise. 
Utilizing pretrained models has become a common practice when developing image classification systems with limited training data~\cite{azizi2021big,he2017fine,SAMBilinear}. These pretrained models provide high-quality feature representations and effectively capture the visual content of images. Nevertheless, it is important to notice that pretrained models are typically trained on image datasets that may differ significantly from the specific fine-grained recognition task at hand. This disparity can potentially introduce bias into the feature representations, where certain visual elements are more prominently represented in the resulting features. However, these visually dominant elements may not necessarily be relevant or useful for the downstream task. It might be expected that these features would either go unused by a classifier trained on the downstream task data or be suppressed during training. However, our initial investigations indicate otherwise.

Figure~\ref{fig:feature_weighting_problem} illustrates a specific scenario to highlight our observations. The upper portion of Figure~\ref{fig:feature_weighting_problem} displays a feature magnitude histogram derived from the pretrained model (i.e., ResNet-50). It is evident that certain feature dimensions exhibit significantly higher magnitudes compared to others. When we employ class activation mapping (CAM)~\cite{zhou2016learning} to investigate the image regions contributing to these features, we discover that some of these features (highlighted in red dashed boxes) do not correspond to regions of interest on the object.
However, when we proceed to train a linear classifier using these features (a.k.a linear probing), we notice that the classifier does not heavily downweight these particular features, as shown in the middle row of Figure~\ref{fig:feature_weighting_problem}. This suggests that despite these features being less relevant to the intended concept for recognition, they can still appear to be discriminative in the context of a small-size training dataset. This situation can mislead the model into relying on these less pertinent features for making final predictions.

\subsection{Feature Magnitude Regularization}
To mitigate the feature magnitude bias inherent in a pre-trained model, we introduce a feature magnitude regularization loss. Initially, we normalize the features using the Softmax operation, which can be expressed as:

\begin{equation}
p_i = \frac{\exp\left (\Psi(I)_i\right )}{\sum_j \exp\left (\Psi(I)_j\right )},
\end{equation} where $\Psi(I)_i$ indicates the value of the $i$-th dimension of the feature $\Psi(I) \in \mathbb{R}^D$.  
This operation produces a pseudo-probability distribution $\mathbf{p} = [p_1, p_2,\cdots,p_D] \in \mathbb{R}^D$. We then apply the negative entropy to form a loss term $\mathscr{L}_\text{fmr}$:

\begin{equation}
\mathscr{L}_\text{fmr} = \lambda \sum_{i=1}^{D}p_i\log(p_i),
\end{equation} where $\lambda$ is a weighting coefficient. 
Please note that when we minimize $\mathscr{L}_\text{fmr}$, we are essentially encouraging the pseudo-distribution $\mathbf{p}$ to closely resemble a uniform distribution. In other words, this optimization aims to ensure that the distribution of magnitudes for the unnormalized features becomes as uniform as possible.

\subsubsection{Dynamic Coefficient Tuning}

The choice of $\lambda$ for $\mathscr{L}_\text{fmr}$ is very important to the successful application of FMR when fine-tuning. A $\lambda$ that is too strong will clobber even useful features, resulting in an uninteresting uniform feature distribution. Moreover, the optimal $\lambda$ value varies from dataset to dataset and throughout the training process itself. In Subsection \ref{subsec:experiment_fmr_weights} below, we explore this in more detail.

To address these challenges, we introduce a dynamic weighting mechanism to adjust the value of $\lambda$ throughout the learning process as follows:

\begin{equation}
\lambda = \beta \times \frac{\mathscr{H}_\text{max} - \mathscr{H}}{\mathscr{H}_\text{max} - \mathscr{H}_\text{init}},
\end{equation} where $\beta$ is a constant, $\mathscr{H}$ is determined by a running average of recent feature vectors' calculated entropies and $\mathscr{H}_\text{max}$ is the maximum possible entropy for a given feature vector size, calculated by:

\begin{equation}
\mathscr{H}_\text{max} = -\log(\frac{1}{D}).
\end{equation}

The initial entropy is obtained before the training begins by:

\begin{equation}
\mathscr{H}_\text{init} = \frac{-1}{N}\sum_{n=1}^N\sum_{d=1}^{D}\mathbf{p}_{n,d}\log(\mathbf{p}_{n,d}).
\end{equation}
where $N$ is the total number of training dataset.
We have set the value of $\beta$ to $50$ in this study, which is optimal in all cases.

The above dynamic weighting scheme can be intuitively understood as follows: $\mathscr{H}_\text{max} - \mathscr{H}_\text{init}$ is the maximal amount of entropy increase we could have during the optimization process and $\mathscr{H}_\text{max} - \mathscr{H}$ denotes the progress that is still to be made toward the target. The effect of this equation is that the value of $\lambda$ is high when there is a large difference between $\mathscr{H}$ and $\mathscr{H}_\text{max}$ and reduces as $\mathscr{H}$ increases. The reduced pressure allows $\mathscr{L}_\text{cls}$ to do its job unmolested.

Through empirical analysis, we observe that this dynamic weighting scheme leads to substantial performance improvements compared to its static counterpart, as detailed in Section \ref{subsec:experiment_fmr_weights}. This finding implies that it may be necessary to apply varying levels of regularization during different stages of optimization. In the initial phases, stronger regularization is required to correct feature magnitude bias. As the feature magnitude distribution becomes more uniform, it becomes unnecessary to further pursue uniformity. 
\section{Experiments}
\label{sec:experiments}

In this section, we evaluate the performance of FMR for three fine-grained visual recognition datasets, as well as on a much larger dataset. The details of these datasets are described in Subsection \ref{subsec:experiment_datasets}. In Subsection \ref{subsec:experiment_results}, we present our obtained performance, along with comparisons to other state-of-the-art methods. In Subsection \ref{subsec:experiment_pretraining_source}, we explore how the pretraining source can affect FMR's usefulness.
In Subsection \ref{subsec:experiment_fmr_weights}, we explore the effect that varying the weighting of the FMR loss has on training outcomes.
Finally, in Subsection \ref{subsec:how_fmr_works}, we discuss how FMR leads to better classification outcomes. The source code behind these experiments is available at
\url{https://github.com/avichapman/feature-magnitude-regularization}.

\subsection{Datasets and Experimental Details}
\label{subsec:experiment_datasets}

\subsubsection{Datasets}

We applied FMR to four popular FGVR datasets: CUB200 \cite{CUB200Birds}, Stanford Cars \cite{StanfordCars}, FGVC-Aircraft \cite{FGVCAircraft} and iNaturalist \cite{iNaturalist}. Due to limited computing resources, we used a subset of iNaturalist consisting of the Order \emph{Passeriformes}. Please see Table~\ref{tab:dataset_details} for details.
To explore the applicability of FMR in low data regimes, we used subsets of the datasets consisting of 15\%, 30\%, 50\% and 100\% of the data.

\begin{table}[]
    \centering
\begin{tabular}{||c c c c||} 
 \hline
 Dataset & Classes & Training Set & Test Set \\ [0.ex] 
 \hline\hline
CUB200~\cite{CUB200Birds} & 200 & 5,994 & 5,794 \\ 
Stanford Cars~\cite{StanfordCars} & 196 & 8,144 & 8,041 \\ 
FGVC-Aircraft~\cite{FGVCAircraft} & 100 & 6,667 & 3,333 \\ 
\makecell{iNaturalist \\ (Passeriformes)}~\cite{iNaturalist} & 678 & 33,900 & 6,780 \\ 
 %iNaturalist & 10,000 & 500,000 & 100,000 \\ 
 \hline
\end{tabular}
\vspace{0.2cm}
\caption{Fine-Grained Visual Recognition Dataset Details}
\label{tab:dataset_details}
\end{table}

\subsubsection{Implementation Details}
Our experiments were conducted using PyTorch, employing a ResNet-50 \cite{Resnet} pretrained on ImageNet \cite{ImageNet} as the backbone network denoted as $\Psi$. Each experimental configuration was repeated three times with and without utilizing the FMR loss. The trade-off parameter $\beta$ for the dynamic loss is set to 50 for all datasets and experiments. 

Following~\cite{SAMBilinear}, the training images were resized to 256$\times$256 pixels and randomly cropped into 224$\times$224 pixel patches. These patches were then subjected to random horizontal flips and RandAugment \cite{cubuk2019randaugment}. We utilized an SGD Optimizer with a batch size of $24$, a learning rate of $0.001$, a momentum of $0.9$, and a weight decay of $0.0001$.
% A plateau scheduler with a patience of $1$ and a learning rate (LR) factor of $0.1$ was employed. The training concluded when the evaluation accuracy remained unchanged for three consecutive plateaus.

During testing, we followed the approach of~\cite{SAMBilinear}, which involved taking five patches and their horizontal reflections, subsequently averaging the predictions obtained from all ten patches.

\subsubsection{Compared Algorithms}
We conducted a performance comparison between FMR and several popular methods for supervised fine-grained visual recognition techniques:
\textbf{SAM - Bilinear} \cite{SAMBilinear}, described above, is the state-of-the-art method for FGVR in the low data regime\footnote{ 
It is important to clarify that our experiment aims to assess learning algorithms in scenarios with limited data availability. As such, we do not engage in direct comparisons with studies that concentrate on the development of network architectures specifically for the FGVR task. Furthermore, \emph{SAM-Bilinear}~\cite{SAMBilinear} has already demonstrated superior performance over various existing FGVR approaches. To maintain a focused and succinct comparison, we have chosen not to include the performance of those additional methods in this study.
}. \textbf{Bilinear Convolutional Neural Networks (B-CNNs)}~\cite{lin2017bilinear} involve passing an image through two Convolutional Neural Networks (CNNs). We compared our results with those reported by Shu \etal \cite{SAMBilinear} who re-implemented this technique using ResNet-50. We also compared against
\textbf{L\textsuperscript{2}-SP}~\cite{L2Sp},
\textbf{DELTA}~\cite{li2020delta} and
Batch Spectral Shrinkage (\textbf{BSS})~\cite{BSS}, which are all described above.
\textbf{Co-Tuning}~\cite{CoTuning} establishes a relationship between the source dataset classes and the target dataset classes. It converts one-hot vectors across the logits for one dataset into probability distributions across the logits for the other dataset and trains both tasks simultaneously. \textbf{MaxEnt}~\cite{dubey2018bentropy} is most similar to our work. We implemented their technique for comparison with ours.
Meanwhile, the performance of naive Fine-Tuning of a pretrained model on the training data is also reported for a reference base and denoted as \textbf{FT Baseline}.

\subsection{Standard FGVR Benchmarks}
\label{subsec:experiment_results}

Our experimental results demonstrate the efficacy of FMR. We first present our results on CUB200, Stanford Cars and FGVC-Aircraft in Table \ref{tab:results_comparison}.

As seen, the proposed methods achieve superior performance compared to existing approaches. For example, at the 15\% training set size for CUB200, FMR achieves a significant lead with an accuracy of 61.30\%, outperforming the next best method (MaxEnt) by nearly \textbf{+7\%}. This trend of superior performance continues across all training set sizes. These results highlight the exceptional ability of the proposed FMR to enhance FGVR with various degrees of data availability, demonstrating its robustness and effectiveness in different training contexts.

\begin{table*}[]
    \centering
    \scalebox{0.8}{
\begin{tabular}{||l|c c c c|c c c c|c c c c|c c c c||} 
 \hline
 & \multicolumn{4}{c|}{CUB200} & \multicolumn{4}{c|}{Stanford Cars} & \multicolumn{4}{c|}{FGVC Aircraft} & \multicolumn{4}{c||}{iNaturalist (Passeriformes)} \\
 Method & 15\% & 30\% & 50\% & 100\% & 15\% & 30\% & 50\% & 100\% & 15\% & 30\% & 50\% & 100\% & 15\% & 30\% & 50\% & 100\% \\ [0.ex] 
 \hline\hline
  L\textsuperscript{2}-SP~\cite{L2Sp} & 45.08 & 57.78 & 69.47 & 78.44 & 36.10 & 60.30 & 75.48 & 86.58 & 39.27 & 57.12 & 67.46 & 80.98 & - & - & - & - \\ 
  DELTA~\cite{li2020delta} & 46.83 & 60.37 & 71.38 & 78.63 & 39.37 & 63.28 & 76.53 & 86.32 & 42.16 & 58.60 & 68.51 & 80.44 & - & - & - & - \\ 
  BSS~\cite{BSS} & 47.74 & 63.38 & 72.56 & 78.85 & 40.57 & 64.13 & 76.78 & 87.63 & 40.41 & 59.23 & 69.19 & 81.48 & - & - & - & - \\ 
  Co-Tuning~\cite{CoTuning} & 52.58 & 66.47 & 74.64 & 81.24 & 46.02 & 69.09 & 80.66 & 89.53 & 44.09 & 61.65 & 72.73 & 83.87 & - & - & - & - \\ 
 \hline
 B-CNNs~\cite{lin2017bilinear} & 49.12 & 63.27 & 73.70 & - & 55.07 & 76.42 & 85.10 & - & 55.06 & 72.12 & 79.93 & - & - & - & - & - \\ 
 SAM bilinear~\cite{SAMBilinear} & 52.35 & 65.19 & 74.54 & - & 57.42 & 77.63 & 85.71 & - & 57.47 & 73.43 & 80.86 & - & - & - & - & - \\ 
 MaxEnt~\cite{dubey2018bentropy} & 54.60 & 67.60 & 75.80 & 80.90 & 60.10 & 77.60 & 85.90 & 91.20 & 54.10 & 71.30 & 78.20 & 86.00 & - & - & - & - \\ 
 \hline
FT Baseline & 50.90 & 64.60 & 74.10 & 81.20 & 52.30 & 73.80 & 83.30 & 90.90 & 53.30 & 70.10 & 77.60 & 86.60 & 15.50 & 29.80 & 39.60 & 50.10 \\ 
\textbf{FMR (Ours)} & \textbf{61.30} & \textbf{71.80} & \textbf{78.20} & \textbf{83.10} & \textbf{64.40} & \textbf{80.40} & \textbf{87.20} & \textbf{91.80} & \textbf{60.20} & \textbf{75.30} & \textbf{81.30} & \textbf{87.30} & \textbf{21.70} & \textbf{35.50} & \textbf{43.90} & \textbf{52.80} \\ 
 \hline
\end{tabular}}
\vspace{0.0cm}
\caption{Classification accuracy (\%) $\uparrow$ on four datasets.}
\label{tab:results_comparison}
% \vspace{-0.5cm}
\end{table*}

We also set out to demonstrate the use of FMR on a much larger dataset. We tested FMR on the subset order \emph{Passeriformes} in the iNaturalist Dataset \cite{iNaturalist} with the same label percentages as we used above. The results can be found in the right-most columns of Table \ref{tab:results_comparison}. FMR again demonstrates superior performance. These results show that FMR works in datasets with larger scale as well. 

 \begin{figure*}
\centering
    \includegraphics[width=\textwidth]{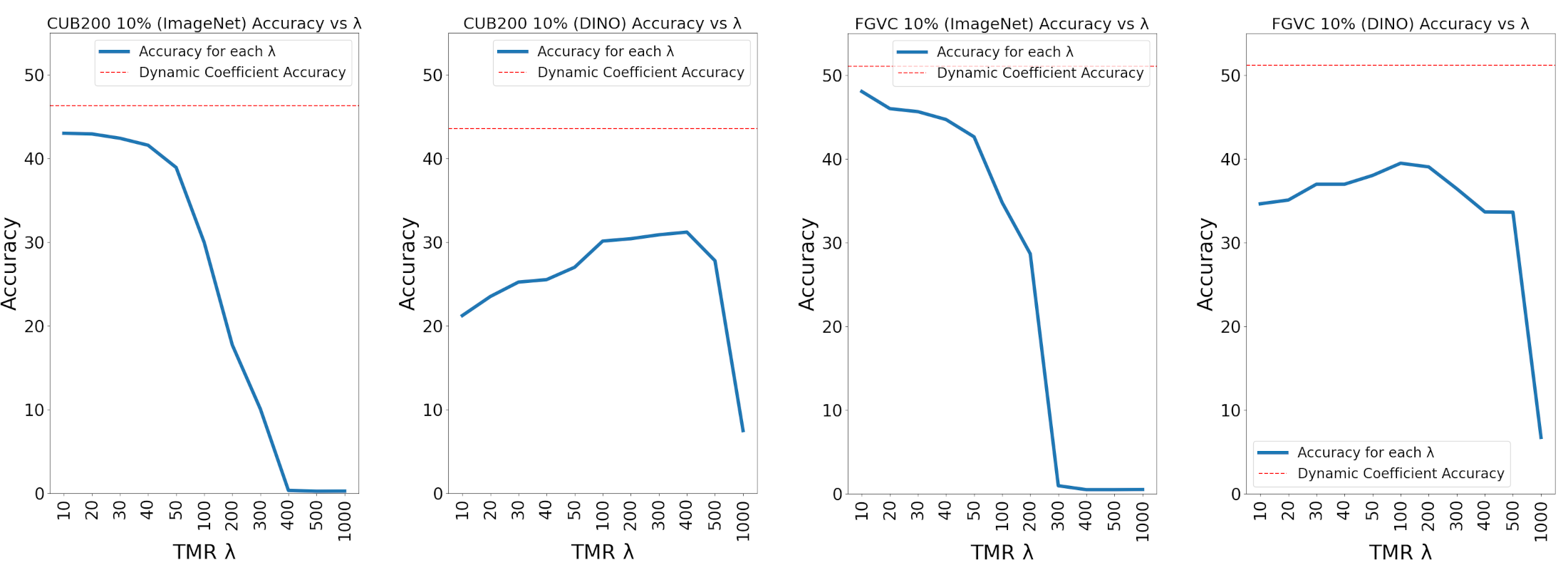}
    \caption{
    Test accuracy (\%) $\uparrow$ comparison of FMR with the proposed dynamic coefficient tuning and with various fixed hyperparameter $\lambda$ on CUB200 10\% and FGVC Aircraft 10\% datasets.
    % FGVC AC and CUB200 pretrained on both ImageNet and DINO have very different starting conditions, making the choice of coefficients potentially difficult. The affect of varying the $\lambda$ value of FMR. The horizontal line shows the accuracy achieved using dynamic coefficient tuning.
    }
    \label{fig:fmr_weight_vs_accuracy}
    \vspace{-0.5cm}
\end{figure*}
\subsection{Ablation Studies}
 
\subsubsection{The Impact of Pretraining Paradigm}
\label{subsec:experiment_pretraining_source}

Given the effectiveness of the proposed FMR in mitigating bias in pretrained models, it's pertinent to explore its impact in relation to the pretraining paradigm used for initial model training.
We conducted two experimental trials to examine this: one where FMR was applied to DINO~\cite{DINO}, a widely recognized self-supervised (unsupervised) method for pretraining a model, and another involving a model with randomly initialized weights, which lacks pretraining and, theoretically, any inherent feature magnitude bias from such a process. To prevent overfitting, especially given our smaller dataset compared to ImageNet, we opted for ResNet-18 for training from scratch.

The results, as illustrated in Table~\ref{tab:dino_results}, are revealing. FMR demonstrated a comparable level of improvement in both supervised and DINO pretrained models, suggesting that the issue of feature magnitude bias might be present even in self-supervised learning models. Intriguingly, when applied to the model trained from scratch, FMR's contribution was minimal, leading to similar outcomes regardless of its use. This aligns with our hypothesis that FMR effectively counters feature magnitude bias inherent in the pretrained models; absent such pretraining, this bias diminishes, rendering FMR less impactful.

\begin{table*}[]
    \centering
    \scalebox{1.0}{
\begin{tabular}{||c|c c|c c|c c||} 
 \hline
 \hline
 & \multicolumn{2}{c|}{ResNet-50 Unsupervised with DINO} & \multicolumn{2}{c|}{ResNet-50 Supervised Pretrained} & \multicolumn{2}{c||}{ResNet-18 With No Pretraining} \\
 \hline
 Data Ratio & FT Baseline & \textbf{FMR (Ours)} & FT Baseline & \textbf{FMR (Ours)} & FT Baseline & \textbf{FMR (Ours)} \\ [0.ex] 
 \hline\hline
15\% & 30.30 ± 0.20 & \textbf{45.30 ± 0.80} & 50.90 ± 0.90 & \textbf{61.30 ± 1.00} & 10.50 ± 0.50\% & \textbf{11.40 ± 0.40\%} \\ 
30\% & 49.30 ± 0.50 & \textbf{63.60 ± 1.20} & 64.60 ± 0.80 & \textbf{71.80 ± 0.30} & 19.10 ± 0.40\% & \textbf{21.30 ± 0.30\%} \\ 
50\% & 63.70 ± 0.40 & \textbf{73.00 ± 0.20} & 74.10 ± 0.40 & \textbf{78.20 ± 0.30} & 31.70 ± 0.60\% & \textbf{33.10 ± 0.70\%} \\ 
100\% & 75.10 ± 1.00 & \textbf{79.30 ± 1.00} & 81.20 ± 0.00 & \textbf{83.10 ± 0.30} & 49.20 ± 1.00\% & \textbf{50.60 ± 0.30\%} \\ 
 \hline\hline
\end{tabular}}
\vspace{0.1cm}
\caption{
Performance comparison of our FMR and fine-tuning baseline on CUB200 dataset under different pretrained methods.
}
\label{tab:dino_results}
\end{table*}

\subsubsection{Dynamic Weighting vs. Static Weighting}
\label{subsec:experiment_fmr_weights}
\begin{figure}
\centering
\includegraphics[width=3.1in]{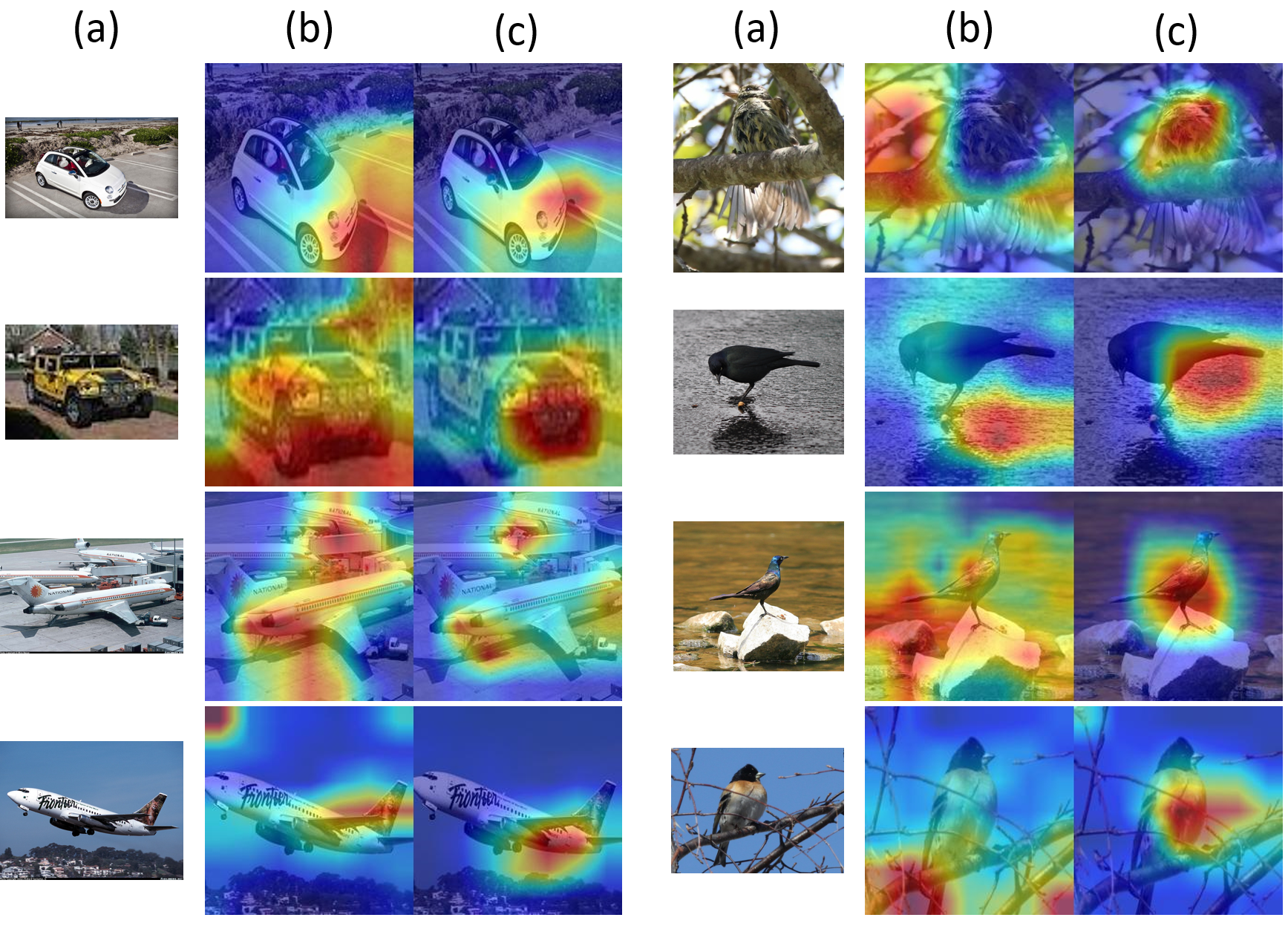}
    \caption{Some sample visualizations of FMR vs. Fine-Tuning Baseline. The FMR results are in column (c), while fine-tuning Baseline results are in column (b). In many cases, the fine-tuning baseline concentrates on incidental details of the background.}
    \label{fig:fmr_cam}
    \vspace{-0.5cm}
\end{figure}

In this section, we explore the impact of employing a dynamic weighting scheme for the proposed feature magnitude regularization module. Specifically, we compare it with an alternative approach that employs a static weighting scheme. We conduct experiments on two datasets, CUB200 and FGVC Aircraft, using two different pretrained backbones. For these experiments, we focus on the scenario where only 10\% of the labeled training data is available. In total, we perform four experiments, varying the FMR weighting coefficient $\lambda$ from 10 to 1000. We record the performance achieved under each $\lambda$ value, generating a performance curve. Additionally, we plot a dash line representing the accuracy obtained by using our proposed dynamic weighting scheme (with $\beta=50$) for comparison. The results are depicted in Figure \ref{fig:fmr_weight_vs_accuracy}.

Figure \ref{fig:fmr_weight_vs_accuracy} clearly illustrates that the choice of $\lambda$ significantly impacts the performance. Interestingly, we observe that irrespective of the static $\lambda$ value chosen, its highest performance consistently falls below that achieved using the dynamic weighting scheme. The performance gap can be substantial, reaching almost 10\% in cases such as when the experiments are conducted with the CUB200 dataset using DINO pretrained ResNet-50 as the backbone. These results provide compelling evidence for the advantages of our proposed dynamic weighting scheme.

\subsection{Analysis of FMR}
\label{subsec:how_fmr_works}

\subsubsection{Encouraging Learning of Generalizable Features}
\label{subsec:encouraging_useful_features}

The motivation behind the proposed method is to address the feature magnitude bias problem commonly encountered in pretrained models. The underlying expectation is that by incorporating the proposed Feature Magnitude Regularization, the model can focus on utilizing more generalizable features while filtering out distracting ones. To quantitatively assess the impact of FMR on the acquisition of more generalizable features, we devise the following experiment.

For both the baseline fine-tuning method and the FMR method, we fix the feature extractor after training on the downstream task dataset. Subsequently, we train two linear classifiers using the features extracted from the backbone: one is trained on the training set, and the other is trained on the testing set. This approach allows us to assess the significance of each feature in distinguishing between data points in the training set and the testing set. Specifically, the weight of the first linear classifier indicates the feature's importance in separating data from the training set, while the weight of the second linear classifier reflects its importance in separating data from the testing set. If a feature exhibits a high degree of generalizability, both classifier weights should have large values, indicating that the feature is deemed important for distinguishing both training and testing data.

\begin{figure}
\centering
    \includegraphics[width=3in]{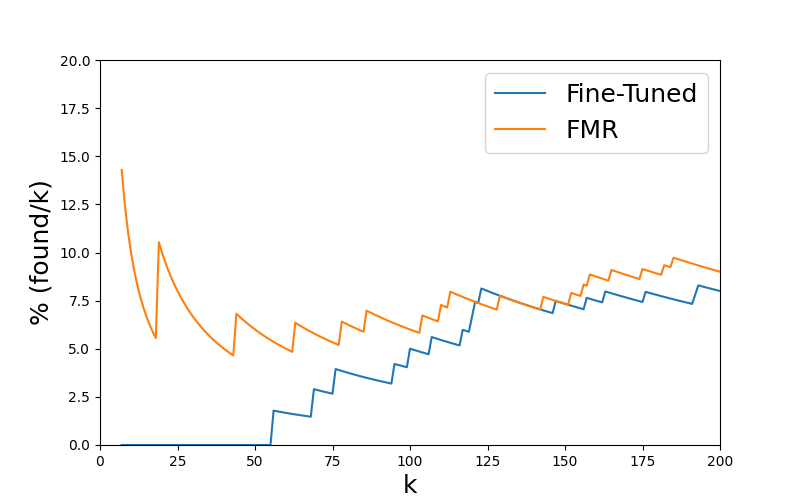}
    \caption{The percentage of top-$k$ weighted features from the training set that also appear in the top-$k$ weighted features of the testing set.}
    \label{fig:good_feature_percentage_in_top_k}
    \vspace{-0.5cm}
\end{figure}

We introduce a measurement that assesses the percentage of top-k weighted features from the training set that also appear in the top-k weighted features of the testing set. A higher percentage indicates the identification of more generalizable features. We present the results for various values of k, and these results are visualized in Figure \ref{fig:good_feature_percentage_in_top_k}. Observing the results, it is evident that the curve associated with the FMR consistently remains above that of the fine-tuning baseline. This trend implies that FMR contributes to the model's ability to recognize more generalizable features

We therefore selected the top-$k$ features by magnitude (averaged across the classes) from the classifier trained on the training set and counted the number $n$ of features that also appeared in the top-$k$ for the testing set. This gave us a percentage - $n/k$. We repeated this exercise with many values of $k$.

The results can be seen in Figure \ref{fig:good_feature_percentage_in_top_k}. This shows that FMR consistently results in the selection of more generalizable features with higher weightings.

\subsubsection{Visualization of the contribution of the top features}
Finally, we employ visualization techniques to gain insights into the image regions that influence the top features learned through different methods. To quantitatively measure the contribution of these top features, we establish the following approach.

For a sample belonging to a specific class, we compute the element-wise product between the feature and its corresponding class weight. This element-wise product reveals the contribution of each dimension to the logit score for that particular class, effectively creating a dimension-wise contribution vector (DCV). Subsequently, we calculate the class-wise mean of the DCV and rank the top-k dimensions within this mean vector. For each sample within the class, we compute the average of the top 5 DCVs to assess the contribution of the top features to the prediction score. We then apply CAM using this average of the top 5 DCVs to identify the corresponding image regions.

Figure \ref{fig:fmr_cam} displays the heat maps visualizing the corresponding image regions obtained from both the fine-tuning baseline approach and our FMR approach. It is evident that our FMR method frequently attend object region whereas the fine-tuned counterpart occasionally directs attention to areas outside of the object.

Combining the outcomes presented in Figure \ref{fig:good_feature_percentage_in_top_k} with the visualizations in Figure \ref{fig:fmr_cam}, these findings provide insights into the characteristics of FMR and its effectiveness in enhancing the generalization performance for fine-grained visual recognition.

%calculate the average value of the feature value in a dimension with its corresponding classifier weight. This value reflects the contribution of the 

%We now present a selection of further visualisations from three datasets.

%Figure \ref{fig:fmr_cam} shows four different ways of visualizing the attended features after standard fine-tuning and FMR fine-tuning:

%\begin{itemize}
%  \item \emph{GradCAM} \cite{GradCAM} is a popular visualization tool for explaining classifier decision making. It uses the gradients in the final layer of the encoder as a weighting value, which is then applied to the activation map output from the encoder.
%  \item \emph{CAM} is the activation map output from the encoder, with each feature multiplied by its weight within the classifier.
%  \item \emph{Features} is a method of our own. We surveyed the entire training set and found the class-wise means of all feature vectors. We then noted the eight most prominent features for each class. When visualising a single sample, we selected those same eight features, given the label for the sample, and added them together to form a heat map.
%  \item \emph{Weighted Features} is similar to \emph{Features}. The difference is that the class-wise mean features have had the classifier's weights applied to them. We only select four features, since fewer feature selections were required to have a unique selection for each class.
%\end{itemize}

%In all four cases, its can be seen that FMR leads to a tighter focus on the discriminative regions.
\section{Conclusions}
\label{sec:conclusions}
In this study, a novel approach named Feature Magnitude Regularization (FMR) was introduced to improve fine-grained image recognition, especially in low-data scenarios. FMR effectively equalizes feature magnitudes, addressing issues arising from dominant features in pre-trained models. This method dynamically adjusts regularization strength based on feature magnitude distribution, leading to more balanced feature representations and improved model performance. Experimental results across various datasets confirmed FMR's superiority over traditional fine-tuning methods, showcasing its potential to enhance image recognition accuracy and generalizability in challenging data-limited environments.

{\small
\bibliographystyle{ieee}
\bibliography{egbib}
}

\end{document}